\documentclass[runningheads]{llncs}
\usepackage[T1]{fontenc}
\usepackage{latexsym}
\usepackage{mathtools}
\usepackage{tabularx}
\usepackage{amsmath}
\usepackage{amssymb}

\begin{document}
\title{Exploring Fake News Detection with Heterogeneous Social Media Context Graphs\thanks{version as accepted at ECIR 2023 with author information, Github link and Acknowledgements added}}
\titlerunning{Fake News Detection}
%
\author{Gregor Donabauer \and Udo Kruschwitz}
%
\authorrunning{Donabauer and Kruschwitz}
%
\institute{Information Science, University of Regensburg, Regensburg, Germany \\ \email{\{gregor.donabauer,udo.kruschwitz\}@ur.de}}
%
\maketitle              
\begin{abstract}
Fake news detection has become a research area that goes way beyond a purely academic interest as it has direct implications on our society as a whole. Recent advances have primarily focused on text-based approaches. However, it has become clear that to be effective one needs to incorporate additional, contextual information such as spreading behaviour of news articles and user interaction patterns on social media. We propose to construct heterogeneous social context graphs around news articles and reformulate the problem as a graph classification task. Exploring the incorporation of different types of information (to get an idea as to what level of social context is most effective) and using different graph neural network architectures indicates that this approach is highly effective with robust results on a common benchmark dataset.

\end{abstract}

\section{Introduction}

Detecting online information disorders is an important step to guarantee free expression and discourse, e.g. \cite{dori_et_al_2021}. A traditional approach to address this problem is manual fact-checking by experts, which is highly labour-intensive, but more recently  automated methods that make use of external knowledge bases and natural language processing (NLP) have been proposed \cite{zhou_zafarani_2020,Guo22Survey}, sometimes as a tool to support the fact-checker \cite{Nakov21Automated}.
Recent studies have demonstrated the growing importance of social media as a source of information for many people \cite{nguyen_et_al_2020,shu_et_al_2017}. In this context it has been pointed out that a range of information types extracted from social media and the relations between such entities are useful for detecting fake news, e.g., \cite{shu_et_al_2019,Mosca21Understanding}. 
Relations between news items and related information signals have been modelled to extract additional features by applying graph neural networks (GNNs) on graph structures (as discussed below), but so far this has typically either been limited to individual social media features or in combination with features that are not taken from social media (which makes it hard to assess the relevance of \emph{social context only}).

To link different types of information they can be modelled as heterogeneous graphs \cite{nguyen_et_al_2020}, and such networks have the capability of introducing multiple types of relations around a news article within a \emph{single data structure}, which allows to see each graph as an independent data point. However it is not clear if such data structures are superior compared to homogeneous graphs that model only one relation type, as spreading and interaction behaviour between true and fake news generally heavily differs on social media \cite{vosoughi_et_al_2018}. 

There is a substantial body of work that demonstrates the use of GNNs, but it remains an open question as to how suitable they are for the use-case of fake news detection with \emph{social media context features only}. It is also unclear which social media signals are most effective here and if there are significant differences between GNN architectures.
To investigate these questions, our contributions can be summarized as follows: \textbf{(1)} We introduce \textbf{HetSMCG}, a methodology to construct extensive \textbf{Het}erogeneous \textbf{S}ocial \textbf{M}edia \textbf{C}ontext \textbf{G}raphs around news articles fusing multiple types of information in a single graph and reformulate fake news detection as a \emph{graph classification problem}; \textbf{(2)} We evaluate a range of experimental settings exploring information sources taken from \emph{social media only} and different GNN architectures to understand how the task can effectively be solved; \textbf{(3)} We compare how performance between heterogeneous multi-relation graphs and homogeneous single-relation graphs differs; \textbf{(4)} We make all our code and in-depth experimental results available to the community to foster reproducibility and sharing of resources.\footnote{\url{https://github.com/doGregor/Graph-FakeNewsNet}}


\section{Related Work}


One way of classifying the different research directions at a high level is to distinguish \emph{content-based} methods that focus on the stories themselves, \emph{context-based}
approaches tapping into, e.g., social media signals and \emph{intervention-based} approaches \cite{Sharma19Combating}. We limit our focus on the context-based paradigm with graph-based representations in our brief contextualization within existing related work.




The majority of context-based approaches for fake news detection use graph structures to model relations, e.g. \cite{shantanu_et_al_2020} created user-article graphs and utilized the neighborhood structure to embed news nodes that can then be classified. However, the number of information types is limited to users and news items. In similar problem formulations, user-post interaction graphs are considered \cite{min_et_al_2022,rode-hasinger-etal-2022-true}. Mehta et al. \cite{mehta_goldwasser_2021} treat fake news detection as a node classification task based on a large graph of users, news and publishers (though they do not model differences in edge types and do not limit information to social media context). In follow up work they also introduce link prediction in this graph structure to find previously unavailable connections \cite{mehta_et_al_2022}. Lu and Li \cite{lu_li_2020} also consider user graphs to extract social interactions using GNNs. In combination with recurrent networks, that leverage retweet patterns, they classify related news articles. Similar approaches that use temporal retweet patterns have been proposed \cite{dou_et_al_2021,song_et_al_2021}. Formulating fake news detection as graph classification has been applied by Dou et al. \cite{dou_et_al_2021}. They calculated user representations by leveraging timeline tweet embeddings and use the outcome as node features to model news propagation paths associated with articles. Another propagation-based approach was introduced by Song et al. \cite{song_et_al_2021}. In their work graph structured temporally evolving retweet patterns have been used to classify whether an article is fake or not. 
Two other approaches left are most similar to our work \cite{nguyen_et_al_2020,het_transformer_2022}. Both are constructing large heterogeneous graphs of articles, users, social media features like tweets, and publishers to classify news nodes as either true or fake (again not limited to social media context). Incoming news nodes need to be integrated into this grown data structure to produce accurate results. With our approach we aim at proposing a method where each article can be seen as an independent data point. 

We conclude that graph structures are a powerful way to include different dimensions of information and thus to improve fake news detection. However, linking many different entity types (and not just one) in independent data structures has barely been investigated yet (and even less if we only consider information types taken from social media in a heterogeneous way) and we aim at contributing to filling this gap with our research.

\section{Methodology and Experiments}

\subsection{Dataset and Graph Construction Method}
\label{subsec:dataset}


The only common benchmark dataset that provides rich social context information in network-like structures is FakeNewsNet \cite{shu_et_al_2020} which is why we adopt it. FakeNewsNet consists of two different datasets, \textit{Politifact} and \textit{GossipCop}.
Both datasets are from different domains (PolitiFact for political news and GossipCop for entertainment stories) which allows to distinguish between domain-specific performances.
Twitter's terms and condition state that tweets and other information cannot be publicly released directly.\footnote{\url{https://developer.twitter.com/en/developer-terms/agreement-and-policy}} We therefore recrawl all data (thereby acquiring a dataset that will slightly differ from what others have used so far).


We construct a heterogeneous social media context graph $\mathcal{G}=(\mathcal{V}, \mathcal{E})$ around each news article. Thus, each snapshot consists of a set of disjoint vertex sets $\mathcal{V} = \mathcal{V}_{N} \cup \mathcal{V}_{T} \cup \mathcal{V}_{U}$ where $\mathcal{V}_{i} \cap \mathcal{V}_{j} = \emptyset, \forall i \neq j$. $\mathcal{V}_{N}$ represents news articles and $|\mathcal{V}_{N}| = 1$ for each graph. Node features for this type of vertex are obtained using BERT-base document embeddings, i.e. $\forall v_{n} \in \mathcal{V}_{N}: v_{n} \in \mathbb{R}^{768}$. All embeddings are generated using flairNLP\footnote{\url{https://github.com/flairNLP/flair}}. Set $\mathcal{V}_{T}$ includes all tweets that refer to the original news article, all tweets retweeting those posts and the latest timeline tweets of each user. Since in our experiments we are incrementally increasing the amount of information types in the heterogeneous graphs, the number of nodes in $\mathcal{V}_{T}$ varies depending on the experimental setup. Node features are a concatenation of textual BERT-base document embeddings, as well as retweet count and favorite count, i.e. $\forall v_{t} \in \mathcal{V}_{T}: v_{t} \in \mathbb{R}^{770}$. Finally, $\mathcal{V}_{U}$ is the set of user nodes where features are a concatenation of BERT-base profile description embeddings, follower count, friends count, favorites count and statuses count, i.e. $\forall v_{u} \in \mathcal{V}_{U}: v_{u} \in \mathbb{R}^{772}$. We also evaluate graphs with text embeddings only and leave out other features (e.g. number of favorite count) in that case.
Edges connecting the nodes are satisfying constraints according to the node types they link together. More specifically, we use at most three types of edges: tweets citing news articles ($(v_{t},\tau_{TN},v_{n}) \in \mathcal{E} \to v_{t} \in \mathcal{V}_{T}, v_{n} \in \mathcal{V}_{N}$), users posting tweets (which also applies to users posting retweets and users posting timeline tweets) ($(v_{u},\tau_{UT},v_{t}) \in \mathcal{E} \to v_{u} \in \mathcal{V}_{U}, v_{t} \in \mathcal{V}_{T}$) and tweets retweeting tweets ($(v_{t},\tau_{TT},u_{t}) \in \mathcal{E} \to v_{t} \in \mathcal{V}_{T}, u_{t} \in \mathcal{V}_{T}$). The graphs fuse all types of information (the news piece as well as a range of different social media context features) without any prior aggregation step.

\subsection{Experimental Setup}

We construct graphs by incrementally increasing the amount of included data as follows: 
\textbf{(1)} We start with only considering tweets related to the news article; 
\textbf{(2)} We add user profiles of the people who posted the tweets to the graphs; 
\textbf{(3)} We include the five latest timeline posts of those users;
\textbf{(4)} We add retweets related to the initial Twitter posts;
\textbf{(5)} We combine all types of information mentioned so far. 
For simplicity we set the number of timeline tweets to five to keep the relation between number of all nodes and number of timeline tweet nodes in the graph manageable. The in-depth exploration of different settings is left as future work. 
To keep the structure of graphs consistent we exclude samples where news data are no longer available.
Graphs with fewer than five vertices per node type are skipped to make sure that at least some social media features are available (other studies sometimes even set higher limits of e.g. a minimum of 15 tweets \cite{lukasik_et_al_2015}). Furthermore, we make all edges undirected to improve message passing during graph convolution. For all experiments we use 5-fold cross-validation and report the average scores as results. We keep the overall number of graphs and the cross-validation split consistent for all setups and datasets to obtain comparable results. The number of graphs per dataset amounts to 483 for PolitiFact (real: 235; fake: 248), 12,214 for GossipCop (real: 10,067; fake: 2,147) and thus 12,697 for the full FakeNewsNet dataset (real: 10,302; fake: 2,395) which consists of a mixture of the two subsets.

We are also flattening all graphs into non-heterogeneous structures to understand how such a setup performs compared to the one described so far. Since homogeneous graphs do not explicitly model disjoint sets of vertices and use a single adjacency matrix, we have to unify all feature dimensions and we do that by either pruning all nodes to a dimension of 768 (only text) or zero-padding all nodes to a dimension of 772.
We classify each graph using two-layer vanilla graph neural networks and evaluate three different types of graph convolution:
(1) GraphSAGE; (2) Graph Attention Convolution (GAT); (3) Heterogeneous Graph Transformer (HGT).
For details on the models we refer to our GitHub.


All models are trained with the same hyperparameter setting to obtain comparable results. We use 20 train epochs, a batch size of 16 and a learning-rate of $8e^{-5}$. As the number of real and fake graphs is unbalanced in terms of GossipCop and the full FakeNewsNet dataset, we use class weights during experiments with those data. Our implementations are based on PyTorch Geometric\footnote{\url{https://github.com/pyg-team/pytorch\_geometric}} \cite{fey_lenssen_2019}.
Training and testing are executed using a single NVIDIA GTX 1080 GPU with an overall graphical memory size of 11GB. Even though we use high-dimensional node features and a large number of graphs during our experiments, training and inference in the described setup only take about 15 minutes on a single GPU (with respect to the highest number of graphs).


\section{Results}

\label{subsec:our_results}

In line with common practice in NLP, we report precision, recall and macro F1 scores for all setups \cite{jurafsky_martin_2021}. Detailed results are reported on our GitHub. As previously mentioned, the values are the average results obtained by 5-fold cross validation.
In general, excluding social features like retweet count and number of followers yields better results than concatenating them with text embedding to initially represent the graph nodes.
We observe the highest scores for Politifact with setup (5) (0.979 macro F1 and accuracy), for GossipCop with setup (3) (0.972 macro F1 and 0.983 accuracy) and the full FakeNewsNet dataset with setup (5) (0.966 macro F1 and 0.979 accuracy), only considering textual features. Interestingly, we get good results for Politifact from the point of adding retweets. Timeline tweets instead are not important. For GossipCop we  observe very similar performance for all setups after once including tweets and users. Here, the other features (retweets and timeline tweets) do not significantly change the model performance. As the full FakeNewsNet dataset mainly consists of GossipCop, we can observe very similar results here. The best performing GNN convolution type is throughout all setups HGT.

The results for the setups with homogeneous graphs are all significantly lower at p < 0.05 compared to their equivalent with heterogeneous graphs. This observation holds true for truncated and padded homogeneous graphs for all three datasets. We demonstrate how modelling disjoint sets of nodes and multiple types of edges improve the representation of social context and lead to overall significantly better fake news detection performance.

For comparison, we use two recently published approaches as strong baselines. 
As we excluded some of the articles from our experiments (due to missing social context information), we rerun both baseline systems with the same data used in our experiments. 
There are many other studies reporting results on FakeNewsNet (e.g. \cite{azri_et_al_2021,dou_et_al_2021,das_et_al_2021,song_et_al_2021}) but we picked the two most competitive, namely:
\textbf{(1) CMTR} \cite{hartl_kruschwitz_2022}: BERT classified texts that were preprocessed using summarization techniques as well as additional social media features like comments. 
\textbf{(2) HetTransformer} \cite{het_transformer_2022}: node classification in a heterogeneous graph featuring user, post and news nodes using an encoder-decoder transformer model.
We use our results obtained by the same setup (HGT and all types of information) for comparison against the baseline scores even if this setup does not always reach the highest
performance in each setup (to avoid cherry-picking). 
Table \ref{table:baselines} reports competitive and robust results on all datasets and
 a new benchmark performance on the Politifact subset (though this not statistically significant). 

\begin{table}[ht!]
\begin{center}
\begin{tabularx}{\textwidth}{X|l|l|l|l|l|l}
      \hline
      \textbf{Approach} & \multicolumn{2}{c|}{\textbf{Politifact}} & \multicolumn{2}{c|}{\textbf{Gossipcop}} & \multicolumn{2}{c}{\textbf{FakeNewsNet}} \\
      \hline
      & \textbf{F1} & \textbf{ACC} & \textbf{F1} & \textbf{ACC} & \textbf{F1} & \textbf{ACC} \\
      \hline
      CMTR \cite{hartl_kruschwitz_2022} & 0.965 & 0.965 & 0.854** & 0.922** & 0.859** & 0.920**\\
      \hline
      HetTransformer \cite{het_transformer_2022} & 0.900** & 0.900** & \textbf{0.994}** & \textbf{0.997}** & \textbf{0.985}** & \textbf{0.991}** \\
      \hline
      HetSMCG (our approach)  & \textbf{0.979} & \textbf{0.979} & 0.969 & 0.982 & 0.966 & 0.979\\
      \hline
\end{tabularx}
\caption{F1 and accuracy scores for \textit{Politifact}, \textit{GossipCop} and the full \textit{FakeNewsNet} dataset. Significant differences to our approach (at $p < 0.05$ and Bonferroni correction) are marked with  with **.}
\label{table:baselines}
\end{center}
\end{table}



\section{Discussion}

An interesting observation is that the results for the Politfact dataset improve when we include retweet data in our network structures. It has already been pointed out that fake news tend to have more retweets than real news \cite{shu_et_al_2020}, and the results provide some support for the utility of this feature.
Moreover, it has also previously been reported that misinformation in general spreads more effectively due to its emotionalizing content \cite{bakir_mcstay_2018}.
Due to the small size of the dataset (real: 239; fake: 261), this could have a large influence on classification performance. The much larger GossipCop and full FakeNewsNet datasets (real: 12,610; fake: 3,185) seems to be less influenced by such information type-specific characteristics. Since we assign retweets to the same node type as article-related tweets and user timeline tweets we address the problem of information type-specific performance gaps. 
An observation that applies to every dataset is that including all types of information leads to the best performing setup. 
For our experiments we also find that there are only marginal differences between the different GNN architectures while HGT gives the highest performance. 



We should also point out some limitations of our work. 
Even though we compare our results against strong, recently published baselines one needs to be careful when reporting and interpreting improvements over existing systems, e.g. \cite{church_kordoni_2022}. Generalisability of insights is a related issue \cite{bowman-2022-dangers}. Providing access to all code and data in our Github repository helps addressing this concern.



\section{Conclusion}


We have demonstrated how including a variety of social media context information can improve fake news detection. By modelling the problem as a graph classification task using heterogeneous graph data structures we achieve competitive results  on a real-world dataset. Incorporating all the contextual information at our disposal leads to the overall best performances. 

\section*{Acknowledgments} 
This work was supported by the project COURAGE: A Social Media Companion Safeguarding and Educating Students funded by the Volkswagen Foundation, grant number 95564.

%
%
%
\bibliographystyle{splncs04}
\bibliography{literature.bib}

\begin{thebibliography}{10}
\providecommand{\url}[1]{\texttt{#1}}
\providecommand{\urlprefix}{URL }
\providecommand{\doi}[1]{https://doi.org/#1}

\bibitem{azri_et_al_2021}
Azri, A., Favre, C., Harbi, N., Darmont, J., No{\^u}s, C.: Monitor: A
  multimodal fusion framework to assess message veracity in social networks.
  In: Bellatreche, L., Dumas, M., Karras, P., Matulevi{\v{c}}ius, R. (eds.)
  Advances in Databases and Information Systems. pp. 73--87. Springer
  International Publishing, Cham (2021)

\bibitem{bakir_mcstay_2018}
Bakir, V., McStay, A.: Fake news and the economy of emotions. Digital
  Journalism  \textbf{6}(2),  154--175 (2018).
  \doi{10.1080/21670811.2017.1345645},
  \url{https://doi.org/10.1080/21670811.2017.1345645}

\bibitem{bowman-2022-dangers}
Bowman, S.: The dangers of underclaiming: Reasons for caution when reporting
  how {NLP} systems fail. In: Proceedings of the 60th Annual Meeting of the
  Association for Computational Linguistics (Volume 1: Long Papers). pp.
  7484--7499. Association for Computational Linguistics, Dublin, Ireland (May
  2022), \url{https://aclanthology.org/2022.acl-long.516}

\bibitem{shantanu_et_al_2020}
Chandra, S., Mishra, P., Yannakoudakis, H., Nimishakavi, M., Saeidi, M.,
  Shutova, E.: Graph-based modeling of online communities for fake news
  detection. CoRR  \textbf{abs/2008.06274} (2020),
  \url{https://arxiv.org/abs/2008.06274}

\bibitem{church_kordoni_2022}
Church, K.W., Kordoni, V.: Emerging trends: Sota-chasing. Natural Language
  Engineering  \textbf{28}(2),  249–269 (2022).
  \doi{10.1017/S1351324922000043}

\bibitem{das_et_al_2021}
Das, S.D., Basak, A., Dutta, S.: A heuristic-driven uncertainty based ensemble
  framework for fake news detection in tweets and news articles. CoRR
  \textbf{abs/2104.01791} (2021), \url{https://arxiv.org/abs/2104.01791}

\bibitem{dori_et_al_2021}
Dori-Hacohen, S., Sung, K., Chou, J., Lustig-Gonzalez, J.: Restoring healthy
  online discourse by detecting and reducing controversy, misinformation, and
  toxicity online. In: Proceedings of the 44th International ACM SIGIR
  Conference on Research and Development in Information Retrieval. p.
  2627–2628. Association for Computing Machinery, New York, NY, USA (2021),
  \url{https://doi.org/10.1145/3404835.3464926}

\bibitem{dou_et_al_2021}
Dou, Y., Shu, K., Xia, C., Yu, P.S., Sun, L.: User preference-aware fake news
  detection. In: Proceedings of the 44th International ACM SIGIR Conference on
  Research and Development in Information Retrieval. p. 2051–2055.
  Association for Computing Machinery, New York, NY, USA (2021),
  \url{https://doi.org/10.1145/3404835.3462990}

\bibitem{fey_lenssen_2019}
Fey, M., Lenssen, J.E.: Fast graph representation learning with {PyTorch
  Geometric}. In: ICLR Workshop on Representation Learning on Graphs and
  Manifolds (2019)

\bibitem{Guo22Survey}
Guo, Z., Schlichtkrull, M., Vlachos, A.: {A Survey on Automated Fact-Checking}.
  Transactions of the Association for Computational Linguistics  \textbf{10},
  178--206 (02 2022). \doi{10.1162/tacl_a_00454}

\bibitem{hartl_kruschwitz_2022}
Hartl, P., Kruschwitz, U.: Applying automatic text summarization for fake news
  detection. In: Proceedings of the Language Resources and Evaluation
  Conference. pp. 2702--2713. European Language Resources Association,
  Marseille, France (June 2022)

\bibitem{jurafsky_martin_2021}
Jurafsky, D., Martin, J.: Speech and Language Processing: An Introduction to
  Natural Language Processing, Computational Linguistics, and Speech
  Recognition. 3rd (draft) edn. (2021),
  \url{https://web.stanford.edu/~jurafsky/slp3/},
  https://web.stanford.edu/\string~jurafsky/slp3/

\bibitem{het_transformer_2022}
Li, T., Sun, Y., Hsu, S.l., Li, Y., Wong, R.C.W.: Fake news detection with
  heterogeneous transformer (2022). \doi{10.48550/ARXIV.2205.03100},
  \url{https://arxiv.org/abs/2205.03100}

\bibitem{lu_li_2020}
Lu, Y.J., Li, C.T.: {GCAN}: Graph-aware co-attention networks for explainable
  fake news detection on social media. In: Proceedings of the 58th Annual
  Meeting of the Association for Computational Linguistics. pp. 505--514.
  Association for Computational Linguistics, Online (Jul 2020).
  \doi{10.18653/v1/2020.acl-main.48},
  \url{https://aclanthology.org/2020.acl-main.48}

\bibitem{lukasik_et_al_2015}
Lukasik, M., Cohn, T., Bontcheva, K.: Point process modelling of rumour
  dynamics in social media. In: Proceedings of the 53rd Annual Meeting of the
  Association for Computational Linguistics and the 7th International Joint
  Conference on Natural Language Processing (Volume 2: Short Papers). pp.
  518--523. Association for Computational Linguistics, Beijing, China (Jul
  2015). \doi{10.3115/v1/P15-2085}, \url{https://aclanthology.org/P15-2085}

\bibitem{mehta_goldwasser_2021}
Mehta, N., Goldwasser, D.: Tackling fake news detection by interactively
  learning representations using graph neural networks. In: Proceedings of the
  First Workshop on Interactive Learning for Natural Language Processing. pp.
  46--53. Association for Computational Linguistics, Online (Aug 2021).
  \doi{10.18653/v1/2021.internlp-1.7},
  \url{https://aclanthology.org/2021.internlp-1.7}

\bibitem{mehta_et_al_2022}
Mehta, N., Pacheco, M., Goldwasser, D.: Tackling fake news detection by
  continually improving social context representations using graph neural
  networks. In: Proceedings of the 60th Annual Meeting of the Association for
  Computational Linguistics (Volume 1: Long Papers). pp. 1363--1380.
  Association for Computational Linguistics, Dublin, Ireland (May 2022),
  \url{https://aclanthology.org/2022.acl-long.97}

\bibitem{min_et_al_2022}
Min, E., Rong, Y., Bian, Y., Xu, T., Zhao, P., Huang, J., Ananiadou, S.:
  Divide-and-conquer: Post-user interaction network for fake news detection on
  social media. In: Proceedings of the ACM Web Conference 2022. p. 1148–1158.
  WWW '22, Association for Computing Machinery, New York, NY, USA (2022).
  \doi{10.1145/3485447.3512163}, \url{https://doi.org/10.1145/3485447.3512163}

\bibitem{Mosca21Understanding}
Mosca, E., Wich, M., Groh, G.: Understanding and interpreting the impact of
  user context in hate speech detection. In: Proceedings of the Ninth
  International Workshop on Natural Language Processing for Social Media. pp.
  91--102. Association for Computational Linguistics, Online (Jun 2021).
  \doi{10.18653/v1/2021.socialnlp-1.8},
  \url{https://aclanthology.org/2021.socialnlp-1.8}

\bibitem{Nakov21Automated}
Nakov, P., Corney, D., Hasanain, M., Alam, F., Elsayed, T., Barrón-Cedeño,
  A., Papotti, P., Shaar, S., Da~San~Martino, G.: {Automated Fact-Checking for
  Assisting Human Fact-Checkers}. In: Zhou, Z.H. (ed.) {Proceedings of the
  Thirtieth International Joint Conference on Artificial Intelligence,
  {IJCAI-21}}. pp. 4551--4558. International Joint Conferences on Artificial
  Intelligence Organization (8 2021). \doi{10.24963/ijcai.2021/619}, survey
  Track

\bibitem{nguyen_et_al_2020}
Nguyen, V.H., Sugiyama, K., Nakov, P., Kan, M.Y.: Fang: Leveraging social
  context for fake news detection using graph representation. In: Proceedings
  of the 29th ACM International Conference on Information \& Knowledge
  Management. p. 1165–1174. CIKM '20, Association for Computing Machinery,
  New York, NY, USA (2020). \doi{10.1145/3340531.3412046},
  \url{https://doi.org/10.1145/3340531.3412046}

\bibitem{rode-hasinger-etal-2022-true}
Rode-Hasinger, S., Kruspe, A., Zhu, X.X.: True or false? detecting false
  information on social media using graph neural networks. In: Proceedings of
  the Eighth Workshop on Noisy User-generated Text (W-NUT 2022). pp. 222--229.
  Association for Computational Linguistics, Gyeongju, Republic of Korea (Oct
  2022), \url{https://aclanthology.org/2022.wnut-1.24}

\bibitem{Sharma19Combating}
Sharma, K., Qian, F., Jiang, H., Ruchansky, N., Zhang, M., Liu, Y.: Combating
  fake news: {A} survey on identification and mitigation techniques. ACM
  Transactions on Intelligent Systems and Technology (TIST)  \textbf{10}(3),
  1--42 (2019), publisher: ACM New York, NY, USA

\bibitem{shu_et_al_2020}
Shu, K., Mahudeswaran, D., Wang, S., Lee, D., Liu, H.: Fakenewsnet: A data
  repository with news content, social context, and spatiotemporal information
  for studying fake news on social media. Big Data  \textbf{8}(3),  171--188
  (2020). \doi{10.1089/big.2020.0062},
  \url{https://doi.org/10.1089/big.2020.0062}, pMID: 32491943

\bibitem{shu_et_al_2017}
Shu, K., Sliva, A., Wang, S., Tang, J., Liu, H.: Fake news detection on social
  media: A data mining perspective. SIGKDD Explor. Newsl.  \textbf{19}(1),
  22–36 (sep 2017). \doi{10.1145/3137597.3137600},
  \url{https://doi.org/10.1145/3137597.3137600}

\bibitem{shu_et_al_2019}
Shu, K., Wang, S., Liu, H.: Beyond news contents: The role of social context
  for fake news detection. In: Proceedings of the Twelfth ACM International
  Conference on Web Search and Data Mining. p. 312–320. WSDM '19, Association
  for Computing Machinery, New York, NY, USA (2019).
  \doi{10.1145/3289600.3290994}, \url{https://doi.org/10.1145/3289600.3290994}

\bibitem{song_et_al_2021}
Song, C., Shu, K., Wu, B.: Temporally evolving graph neural network for fake
  news detection. Information Processing \& Management  \textbf{58}(6),  102712
  (2021). \doi{https://doi.org/10.1016/j.ipm.2021.102712},
  \url{https://www.sciencedirect.com/science/article/pii/S0306457321001965}

\bibitem{vosoughi_et_al_2018}
Vosoughi, S., Roy, D., Aral, S.: The spread of true and false news online.
  Science  \textbf{359}(6380),  1146--1151 (2018).
  \doi{10.1126/science.aap9559},
  \url{https://www.science.org/doi/abs/10.1126/science.aap9559}

\bibitem{zhou_zafarani_2020}
Zhou, X., Zafarani, R.: A survey of fake news: Fundamental theories, detection
  methods, and opportunities. ACM Comput. Surv.  \textbf{53}(5) (sep 2020).
  \doi{10.1145/3395046}, \url{https://doi.org/10.1145/3395046}

\end{thebibliography}
%




\end{document}